\documentclass[letterpaper, 10 pt, conference]{ieeeconf}  

\IEEEoverridecommandlockouts                              
\usepackage{times}
\usepackage{graphicx}
\usepackage{amsmath,amssymb,amsopn,amstext,amsfonts}
\usepackage{cancel}
\usepackage[space]{cite}
\usepackage{pdfsync}
\usepackage{balance}
\usepackage{color}
\usepackage{mathtools}
\usepackage{algpseudocode}
\usepackage{algorithm}
\usepackage{bm}

\usepackage{diagbox}
\usepackage{float}
\usepackage{epstopdf}
\usepackage{pifont}
\usepackage{fixltx2e}
\usepackage{amsmath}
\usepackage{multirow}
\usepackage{url}
\usepackage{verbatim}
\usepackage{amssymb}
\makeatletter
\let\NAT@parse\undefined
\makeatother
\usepackage[linkcolor=black,citecolor=black,urlcolor=black,colorlinks=TRUE]{hyperref}
\usepackage{booktabs}
\usepackage{graphicx}
\usepackage{subcaption}
\usepackage{gensymb}
\usepackage{stfloats}
\usepackage{amsmath}
\usepackage{nccmath}

\graphicspath{{./FIGURES/}}
\DeclareGraphicsExtensions{.png,.jpg,.eps,.pdf}

\title{\LARGE \bf
	Shape-Adaptive Planning and Control for a Deformable Quadrotor
}

\author{Yuze Wu\textsuperscript{1,2,$\dagger$}, Zhichao Han\textsuperscript{1,2,$\dagger$}, Xuankang Wu\textsuperscript{2,3,$\dagger$}, Yuan Zhou\textsuperscript{1,2}, \\ Junjie Wang\textsuperscript{1,2}, Zheng Fang\textsuperscript{3}, and Fei Gao\textsuperscript{1,2}
	\thanks{This work was supported by the National Key R\&D Program of China under grant no. 2023YFB4706600. (\emph{Corresponding author: Fei Gao})} 
	\thanks{$^\dagger$These authors contributed equally to this work.}
	\thanks{\textsuperscript{1}State Key Laboratory of Industrial Control Technology, Zhejiang University, Hangzhou 310027, China.}
	\thanks{\textsuperscript{2}Huzhou Institute, Zhejiang University, Huzhou 313000, China.}	
	\thanks{\textsuperscript{3}Faculty of Robot Science and Engineering, Northeastern University, Shenyang 110819, China.}	
	\thanks{E-mail:{\tt\small \{wuyuze000, fgaoaa\}@zju.edu.cn}}
}

\begin{document}

	\makeatletter
	\let\@oldmaketitle\@maketitle
	\renewcommand{\@maketitle}{\@oldmaketitle
		\begin{center}
			\includegraphics[width=1.0\linewidth]{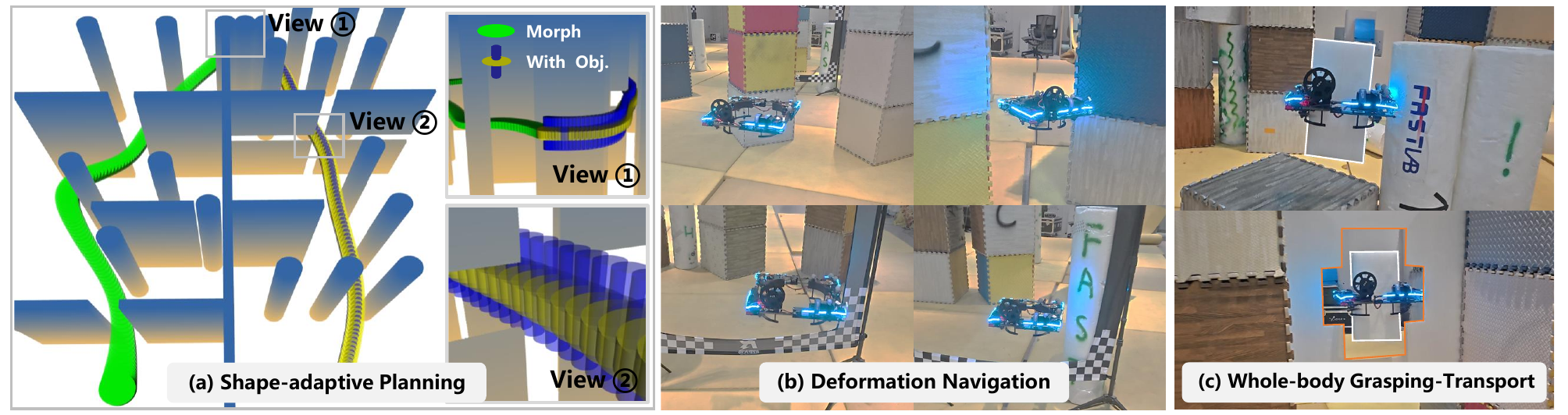}
		\end{center}
		\captionsetup{font={small}}
		\captionof{figure}{
			\label{fig:top}
			(a) Shape-adaptive motion planning for deformable traversal and whole-body grasping-transport. (b) Highlight of autonomous deformation enabling traversal through confined areas. (c) Highlight of autonomous grasping-transport through a narrow cross-shaped gap.
			\vspace{-0.4cm}
		}
	}
	\makeatother
	\maketitle
	\setcounter{figure}{1}
	\thispagestyle{empty}
	\pagestyle{empty}

	\begin{abstract}
		
		Drones have become essential in various applications, but conventional quadrotors face limitations in confined spaces and complex tasks. Deformable drones, which can adapt their shape in real-time, offer a promising solution to overcome these challenges, while also enhancing maneuverability and enabling novel tasks like object grasping. This paper presents a novel approach to autonomous motion planning and control for deformable quadrotors. We introduce a shape-adaptive trajectory planner that incorporates deformation dynamics into path generation, using a scalable kinodynamic A* search to handle deformation parameters in complex environments. The backend spatio-temporal optimization is capable of generating optimally smooth trajectories that incorporate shape deformation.
		Additionally, we propose an enhanced control strategy that compensates for external forces and torque disturbances, achieving a 37.3\% reduction in trajectory tracking error compared to our previous work. Our approach is validated through simulations and real-world experiments, demonstrating its effectiveness in narrow-gap traversal and multi-modal deformable tasks.
		
	\end{abstract}

	\section{INTRODUCTION}
	
	Unmanned Aerial Vehicles (UAVs) are increasingly vital in various scenarios, ranging from dynamic aerial cinematography to intricate tunnel exploration. However, conventional quadrotors, constrained by their fixed physical dimensions, exhibit inherent limitations. Larger drones often encounter significant maneuverability challenges within confined spaces, hindering navigation through narrow passages. Conversely, smaller drones, while offering improved accessibility, frequently compromise flight endurance and payload capacity, and demonstrate reduced robustness against external disturbances like wind. To overcome these limitations, the concept of deformable drones has emerged as a promising paradigm shift \cite{2019Falanga,2017Desbiez,2018Moju}. These unique platforms possess active morphing capabilities, enabling real-time adjustments to their physical dimensions.  This dynamic adaptability not only significantly enhances navigation within complex and spatially restricted environments but also unlocks unprecedented operational potentials beyond traditional UAVs, including the grasping and transport of objects \cite{2019Falanga,2017Moju,Bucki2021DesignAC}, thereby expanding the applicability across diverse human-interactive domains, for example, indoor logistics.
	
	In our prior work \cite{wu2023ring}, we introduced a lightweight deformable quadrotor hardware platform, characterized by a single actuator enabling dual-degree-of-freedom (DOF) deformation. This platform was experimentally validated for its capacity to perform grasping-transport tasks.  Nonetheless, it inherently lacks autonomous motion planning capabilities. The execution of complex operational procedures thus necessitates reliance on either direct human teleoperation or pre-programmed sequences. This dependency naturally restricts its autonomous application and incurs significant human resource expenditure.
	Traditional motion planning algorithms, however, are fundamentally ill-suited for direct application to deformable UAVs. These methodologies typically disregard the inherent DOF associated with shape deformation, often approximating the robot's spatial footprint as a fixed-size sphere~\cite{tordesillas2021mader,zhang2023model}. Such simplifications inevitably sacrifice solution space and compromise optimality. Furthermore, in intricate grasping-transport scenarios, conventional approaches frequently neglect to incorporate the shape of the grasped object into obstacle avoidance considerations, potentially leading to collisions. Upon trajectory generation, effective trajectory tracking necessitates a robust controller.  Unfortunately, morphing and grasping tasks often cause dynamic alterations in system dynamics, varying loads, and unmodeled disturbances, severely degrading control performance.

	\begin{figure*}[t]
		\centering
		\includegraphics[width=1.0\linewidth]{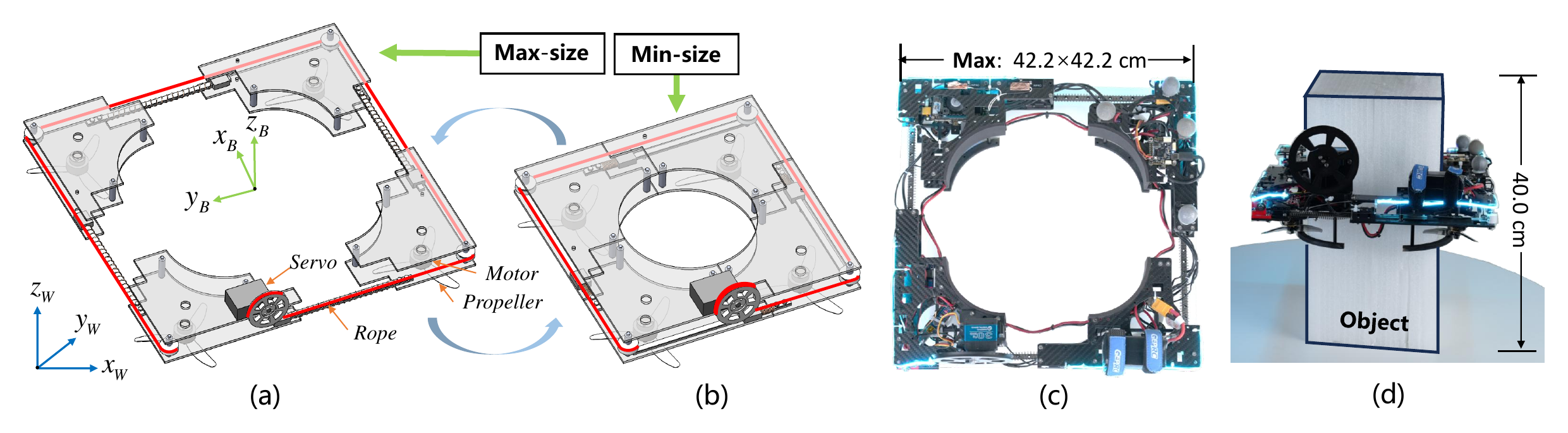}
		\captionsetup{font={small}}
		\caption{(a-b) Mechanical schematics of the morphing quadrotor named RR: the robot utilizes a single motor and tendon-driven system for dynamic switching between maximum and minimum sizes. (c) RR's hardware platform. (d) RR is capable of whole-body object grasping.}
		\label{pic:pltform}
		\vspace{-0.5cm}
	\end{figure*}

	To address these challenges, we introduce a novel shape-adaptive motion planner that generates trajectories inherently accounting for shape deformation in response to environments (Fig.~\ref{fig:top}(b)), following a hierarchical framework. In the front-end, a scalable kinodynamic A* path search extends the robot's state space to incorporate deformation parameters and performs a discrete search to construct a search tree from the initial state, ultimately yielding a path with shape transformations. Subsequently, this path initializes the back-end trajectory optimization module, which formulates a nonlinear optimization problem leveraging a compact piecewise polynomial representation in both Cartesian and deformation spaces to ensure robust convergence to optimal smooth trajectories in continuous time-space (Fig.~\ref{fig:top}(a)).  
	Notably, our formulation naturally possesses extensibility and unity, readily accommodating the volume of grasped objects into obstacle avoidance considerations (Fig.~\ref{fig:top}(c)).
	Furthermore, to enhance trajectory tracking precision during deformation maneuvers, we propose an enhanced control strategy.  This strategy adopts a nonlinear model predictive controller (NMPC) as the baseline, and integrates real-time external force estimation, compensated within the thrust controller, with an incremental nonlinear dynamic inversion (INDI) algorithm to effectively counteract external torque disturbances.  Experimental evaluations demonstrate that this enhanced control strategy achieves a 37.3\% reduction in trajectory tracking error compared to our previous work \cite{wu2023ring}.
	Consequently, our contributions are summarized as follows:
	\begin{itemize}	
		\item [1)]
		We develop a shape-adaptive trajectory planner to co-optimize both flight trajectory and morphing states, suitable for the quadrotor's deformable flight in complex and narrow environments.
		
		\item [2)]
		We introduce an enhanced controller with force and torque compensation to reduce the tracking error for the multi-modal deformable tasks.
		
		\item [3)] 
		Extensive simulations and real-world experiments validate the performance of shape-adaptive motion planning and enhanced control.
	\end{itemize}

	\section{Related Work} 
	\label{sec:related_works}
	
	\subsection{Motion Planning with Morphing}
	
	Traditional motion planners~\cite{ren2022bubble} typically employ gradient-based methods to generate feasible trajectories, often employing simplified robot models such as spheres or point masses, which results in overly conservative solutions.
	Alternatively, the recent work~\cite{han2021fast,wu2024whole} plans in $\mathbb{SE}$(3) for full-shape obstacle avoidance, enabling aggressive maneuvers in confined spaces; however, it demands substantial control effort.  Deformable UAVs, in contrast, offer a more natural and efficient approach to navigating narrow spaces by dynamically reducing dimensions.
	Nevertheless, research on autonomous morphing motion planning remains insufficient in the industry. Bucki et al. \cite{Bucki2021DesignAC} rely on ballistic motion to achieve morphing traversal, but their trajectories are typically predefined. A similar approach is employed in Zhao et al.'s "aerial robot dragon" \cite{2018Moju}.
	Cui et al.~\cite{cui2024motion} proposed a full-shape trajectory planning framework for morphing drones based on rectangular envelopes.
	However, this work was not validated in large-scale, dense environments.  Moreover, from a technical standpoint, it lacks a shape-aware front-end module. Consequently, optimization is constrained to topologies from manual assignment or point-mass graph searches, inherently leading to suboptimal trajectories. 
	Furthermore, this approach narrowly focuses on deformation for obstacle avoidance, overlooking the consequential effects of varying shapes on the fundamental flight dynamics such as energy expenditure, demonstrably deviating from optimal solutions.
	
	\subsection{Flight Control with Morphing}
	
	Morphing drones face control challenges such as morphing interference, model mismatch, and external disturbances during the deformation process, which can degrade trajectory tracking performance. This issue becomes even more critical in high-precision tracking scenarios, such as gap drilling, where accurate control is essential. In existing studies, Falanga et al. \cite{2019Falanga} used a linear quadratic regulator (LQR) controller for deformation flight, Derrouaou et al. \cite{derrouaoui2021pso} and Kim et al. \cite{kim2021morphing} employed linear PID controllers for flying and grasping. Hu et al. \cite{2021Hu} achieved stable hovering using reinforcement learning.  However, these studies did not account for aerodynamic disturbances. To address this, Cui et al. \cite{cui2024motion} proposed a nonlinear attitude controller with aerodynamic drag compensation. However, these methods have not effectively addressed model mismatch and additional disturbances. This study integrates an external force observation feedback mechanism into the position loop and introduces an INDI torque compensation mechanism in the attitude loop, enabling active compensation for morphing disturbances, model deviations, and external disturbances.

	\section{DYNAMIC MODEL}
	\label{sec:DYNAMIC_MODEL}
	In previous work, we proposed the deformable and grasping integrated quadrotor named RR (Fig.~\ref{pic:pltform}(c)). The robot adopts the novel ring-shaped configuration consisting primarily of four parts connected to adjacent parts. Based on this tandem structure, distances between adjacent parts can be changed synchronously using only one actuator and tendon drive mechanism, achieving two dimensions of size reduction (Fig.~\ref{pic:pltform}(a-b)).
	Moreover, RR possesses a novel whole-body aerial grasping and transport capability that fits various shapes of objects (Fig.~\ref{pic:pltform}(d)), which can be easily used for disaster relief, package delivery, and other fields. In this work, we modify the mechanical structure of RR to enhance morphing performance. Now it has a maximum size of 42.2$\times$42.2 cm, and can be reduced by a maximum of 37.8\%, resulting in a minimum size of 26.2$\times$26.2 cm.
	
	Then, we model the dynamics of the quadrotor. As shown in Fig.~\ref{pic:pltform}(a), we define the world coordinate system \(\{\mathbf{x}_W, \mathbf{y}_W, \mathbf{z}_W\}\) and the body coordinate system \(\{\mathbf{x}_B, \mathbf{y}_B, \mathbf{z}_B\}\).
	Let $\boldsymbol{p}_W$ = $(\emph{p}_x, \emph{p}_y,\emph{p}_z)^{\rm T}$, $\boldsymbol{q}_W$ = $(\emph{q}_x, \emph{q}_w, \emph{q}_y,\emph{q}_z)^{\rm T}$ and $\boldsymbol{v}_W$ = $(\emph{v}_x, \emph{v}_y,\emph{v}_z)^{\rm T}$ be the position, the orientation and the linear velocity of the quadrotor expressed in the world frame. Additionally, let  $\boldsymbol{\omega}_B$ = $({\omega}_x, {\omega}_y, {\omega}_z)^{\rm T}$ its angular velocity in the body frame. 
	
	The quadrotor model is established using 6-DoF rigid body kinematic and dynamic equations. For translational dynamics, we have
	\begin{equation}
		\begin{aligned} 
			& \dot{\boldsymbol{p}}_W = {\boldsymbol{v}}_W, \\
			& \dot{\boldsymbol{v}}_W = (F\boldsymbol{z}_B + \boldsymbol{f}_{ext})/m + \boldsymbol{g},
			\label{con:trans_dynamics}
		\end{aligned}
	\end{equation}
	where \emph{F} and \emph{m} are the collective thrust and total mass respectively; $\boldsymbol{z}_B$ is the Z axis of the body frame expressed in the world frame; \emph{\textbf{g}} = $[0, 0, -g]^{\rm T}$ is the gravitational vector; $\boldsymbol{f}_{ext}$ indicates the external aerodynamic drag force.
	
	The rotational kinematic and dynamic equations are expressed as
	\begin{equation}
		\begin{aligned} 
			& \dot{\boldsymbol{q}}_W = \frac{1}{2}(\begin{bmatrix}
				0 \\  
				\bm{\omega}_B \\  
			\end{bmatrix}_{\times}) \cdot \boldsymbol{q}_W,  \\
			& \dot{\bm{\omega}}_B = \boldsymbol{J}^{-1}(\bm{\tau} - \bm{\omega}_B\times\boldsymbol{J}\bm{\omega}_B + \bm{\tau}_{ext}),
			\label{con:rota_dynamics}
		\end{aligned}
	\end{equation}
	where $[\cdot]_{\times}$ is the skew-symmetric matrix; $\boldsymbol{\tau}$ and $\boldsymbol{J}$  are the total torque and inertia tensor matrix respectively; $\boldsymbol{\tau}_{ext}$ is external torque disturbances.
	
	Let $k_t$ ,$k_c$ be the thrust coefficient and torque coefficient of the $j$-th motor. $\Omega_j$ and $\boldsymbol{l}_j = [l_{x_j},l_{y_j},l_{z_j}]^{\rm T}$ are the rotational speed of the $j$-th motor and its position in the body frame. The collective thrust $\emph{F}$ and torque $\boldsymbol{\tau}$ are generated by the actuators, expressed by
	\begin{equation}
		\begin{aligned} 
			\begin{bmatrix}
				F \\  
				\boldsymbol{\tau}\\ 
			\end{bmatrix} = \boldsymbol{H}_k\boldsymbol{t},
		\end{aligned}
	\end{equation}
	where $\boldsymbol{t}=[k_t\Omega^2_1, k_t\Omega^2_2, k_t\Omega^2_3, k_t\Omega^2_4]^{\rm T}$ represents the thrust generated by each rotor, $\boldsymbol{H}_k$ is the time-variant control allocation matrix while the quadrotor morphing.

	\begin{figure}[t]
		\centering
		\includegraphics[width=0.9 \linewidth]{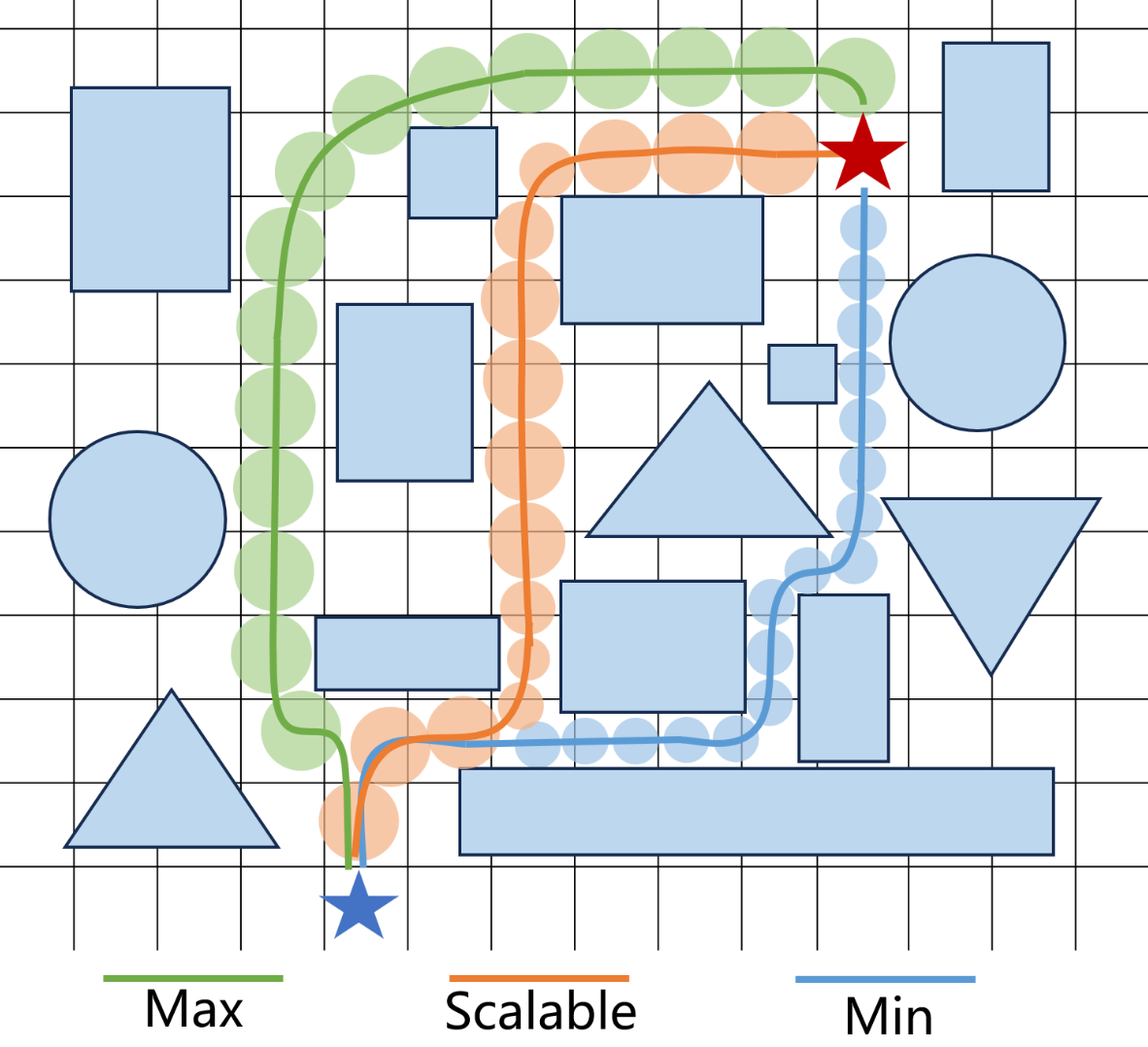}
		\captionsetup{font={small}}
		\caption{
			Comparative schematic diagram of front-end kinodynamic A* pathfinding with max size, min size, and scalable size.
		}
		\label{pic:astar}
		\vspace{-0.9cm}
	\end{figure}

	\section{METHODOLOGY}
	\subsection{Scalable Kinodynamic A* Path Search} 
	\label{sec:MR_A_star}
	
	The traditional kinodynamic A* algorithm treats robots as point masses or fixed-size entities when searching for traversable paths, which is not suitable for multi-modal deformable UAVs. 
	In this study, we propose a scalable kinodynamic A* pathfinding algorithm that explicitly incorporates radius deformation into the state space representation, thereby balancing obstacle avoidance and endurance, as shown in Fig.~\ref{pic:astar}. The state space 
	$\boldsymbol X$ = $[\boldsymbol p_W^{\rm T},\emph{r},\boldsymbol v_W^{\rm T}, \emph{v}_r]^{\rm T}$
	is formally defined as a composite vector comprising the centroid position $\boldsymbol p_W$, deformable radius $\emph{r}$, and their first-order time derivatives. The control inputs $\boldsymbol u$ are specified as acceleration parameters governing system dynamics.
	The state transition mechanism is mathematically formulated through the discrete-time propagation equation:
	\begin{align}
		\boldsymbol X_{k+1} &= \boldsymbol A \boldsymbol X_k + \boldsymbol B \boldsymbol u,\\    
		\boldsymbol A &= \begin{bmatrix} 
			\boldsymbol I_4 & \boldsymbol I_4 \Delta T \\ 
			\boldsymbol 0 & \boldsymbol I_4
		\end{bmatrix}, \quad
		\boldsymbol B = \begin{bmatrix} 
			\frac{1}{2} \boldsymbol I_4 (\Delta T)^2 \\ 
			\boldsymbol I_4 \Delta T 
		\end{bmatrix},
	\end{align}
	where $\Delta T$ denotes the temporal interval allocated for each node expansion in the search process. During tree exploration, our algorithm systematically generates successor states $\boldsymbol X_{k+1}$ by sampling discretized control inputs $\boldsymbol u\in[-\boldsymbol u_{\max}, \boldsymbol u_{\max}]$ and temporal intervals $\Delta T\in[\Delta T_{\min}, \Delta T_{\max}]$ from admissible sets. 
	We incorporate Euclidean Signed Distance Field (ESDF) field to rigorously enforce safety constraints during path exploration:
	\begin{align}
		D_{margin}-\mathcal{D}_\mathcal{S}(\boldsymbol p_W, \emph{r})\leq 0,
		\label{eq:safepath}
	\end{align}
	$\mathcal{D}_\mathcal{S}$ refers to the obstacle clearance metric accounting for the robot's complete geometry $\mathcal{S}$.
	Taking grasping-transport tasks as an example, the robot is modeled as a cylinder with radius $\emph{r}$ and fixed height $\emph{h}$ in the pre-grasping stage; in the post-grasping stage, the geometry of the grasped object is additionally considered in the collision model.
	In practical implementation, the safety constraint Eq. (\ref{eq:safepath}) is enforced by discretizing the robot's complete geometry into sampled points, where all points are required to maintain signed distances above the specified threshold $D_{margin}$,
	with its detailed mathematical formulation provided in the next section.
	Beyond obstacle avoidance, nodes exceeding predefined velocity or deformation rate bounds are systematically pruned to ensure the physical feasibility of generated paths.
	
	The findings~\cite{2019Falanga,wu2023ring} indicate that smaller dimensions of morphing quadrotors frequently lead to blade overlap, blade interference and airflow blockage, which reduce flight endurance. Furthermore, reduced dimensions also imply shorter moment arms, which can negatively impact control precision and disturbance rejection capabilities.  Consequently, unless dictated by obstacle avoidance requirements, it is generally preferable to maximize the robot's dimensions for smoother and more energy-efficient flight.  Therefore, our composite cost metric $\emph{g}_c$ for motion primitives during node expansion integrates a second-order radius regularization term, and conventional metrics for path smoothness and flight time:
	\begin{align}
		\emph{g}_c = ||\boldsymbol{u}||_2^2\Delta T + a(\frac{\emph{r}-\emph{r}_{\max}}{\emph{r}_{\max}})^2\Delta T 
		+ w_{T}\Delta T,
	\end{align}
	where $a$ and $w_{T}$ are user-defined weights.
	Following conventional A* principles, an admissible and informative heuristic is critical for accelerating the search process. Inspired by the work~\cite{zhou2019robust}, we employ cubic splines to analytically connect the current expanded node state with the goal state, where the spline's cost metric is adopted as the heuristic function.

	\subsection{Shape-Adaptive Trajectory Optimization} 
	\label{sec:EA_planning}
	
	While the front-end search yields collision-free paths, the low-dimensional and discretized nature of the search space frequently results in insufficient trajectory quality for direct robotic execution. To address this limitation, we introduce a subsequent trajectory optimization framework
	in continuous space-time space that utilizes the initial path as a warm-start, generating refined trajectories through systematic consideration of higher-order kinematic constraints and smoothness requirements.  
	Diverging from conventional approaches that model UAVs as point masses or fixed-radius spheres, our novel formulation incorporates the co-optimization of centroid motion and deformation parameters.   
	Given control effort degree $s$, the composite trajectory in the flat space $\bm\sigma(t) = {[\boldsymbol{p}_W^{\rm T}, \emph{r}]}^{\rm T}(t) : [0,T]$ is represented using $\mathcal{M}$-wise polynomial functions 
	with degree $N=2s-1$ that jointly represent geometric deformation radius $\emph{r}$ and centroid motion $\boldsymbol{p}_W$, providing a compact representation while structurally enforcing $\mathcal{C}^2$ continuity across all motion states.
	Under this formulation, the trajectory is parameterized by polynomial coefficients $\mathbf{c} = [\bm{c}_1^{\rm T},...,\bm{c}_i^{\rm T}, \bm{c}_{\mathcal{M}}^{\rm T}]^{\rm T} \in \mathbb{R}^{2\mathcal{M}s\times 4}$ and temporal intervals 
	$\bm T = [T1,...,T_i,...,T_{\mathcal{M}}]^{\rm T} \in \mathbb{R}^{\mathcal{M}}$
	for each piece. The $i$-th trajectory piece $\bm \sigma_i$ is consequently expressed as:
	\begin{align}
		\bm \sigma_i(t) &= \bm{c}_i^{\rm T}\bm\beta(t),\\
		\bm \beta(t) &= [1,t,t^2,...,t^N]^{\rm T},\\
		\forall t &\in [0, T_i], \forall i \in \{1,...,\mathcal{M}\}.
	\end{align}
	
	Then, the complete trajectory representation $\bm\sigma(t) : [0,T]$ is formulated:
	\begin{small}
		\begin{align}
			\bm\sigma(t) &= \bm\sigma_i(t-\sum_{k=0}^{i-1}T_k),\\
			\forall i &\in \{1,...,\mathcal{M}\}, \forall t \in [\sum_{k=0}^{i-1}T_k, \sum_{k=0}^{i}T_k].
		\end{align}
	\end{small}
	The trajectory planning problem is formulated as a nonlinear constrained optimization that simultaneously minimizes the position control effort (PCE), the radius control effort (RCE), the second-order radius regularization term (SORR) and the flight time cost, subject to obstacle avoidance constraints and dynamic feasibility requirements:
	\begin{align}
		\min_{\mathbf{c}, \bm T} J &= 
		\int_{0}^{T} \boldsymbol{p}_W^{(s)}(t)^{\rm T} \boldsymbol{p}_W^{(s)}(t) 
		+ (\emph{r}^{(s)}(t))^2 \nonumber \\
		&+ a(\frac{\emph{r}(t)-\emph{r}_{\max}}{\emph{r}_{\max}
		})^2dt + w_TT \label{eq:opt}\\
		s.t. & \bm \sigma^{[s-1]}(0) = \bm \sigma_0, \bm \sigma^{[s-1]}(T) = \bm \sigma_f
		\label{eq:bdcon}
		\\
		&||\boldsymbol{p}_W^{(1)}(t)||_2^2 \leq v_{\max}^2, ||\boldsymbol{p}_W^{(2)}(t)||_2^2 \leq a_{\max}^2,\label{eq:va}\\
		&(\emph{r}^{(1)}(t))^2 \leq \omega_{\max}^2, (\emph{r}^{(2)}(t))^2 \leq \alpha_{\max}^2,\label{eq:omegaalpha}\\
		&D_{margin}-\mathcal{D}_\mathcal{S}(\boldsymbol p_W(t), \emph{r}(t))\leq 0,\label{eq:safe}\\
		&\bm \sigma^{[\widetilde{d}]}_i(T_i) = \bm \sigma^{[\widetilde{d}]}_{i+1}(0), 
		\label{eq:continus}
		\\
		&\forall t \in [0,T], \forall i \in \{1,...,\mathcal{M}\}.
	\end{align}
	Eq. (\ref{eq:bdcon}) enforces boundary constraints by ensuring strict adherence of the trajectory's initial and terminal states to prescribed conditions, defined as the robot's position, deformable radius, and their time derivatives up to the $(s-1)$-th order.
	Eq. (\ref{eq:va}) and Eq. (\ref{eq:omegaalpha}) enforce dynamic feasibility constraints by restricting the robot's maximum velocity, acceleration, deformation velocity, and deformation acceleration to prescribed thresholds.
	Prior to object grasping, the robotic system is abstracted as a cylindrical volume with configurable radius, where 
	the full-shape safety condition Eq. (\ref{eq:safe}) 
	is approximately enforced through discretized sampling of surface points $\boldsymbol q_B$:
	\begin{align}
		&D_{margin}-\mathcal{D}_\mathcal{S}(\boldsymbol p_W, \emph{r})\leq 0 \sim \nonumber\\
		&D_{margin}-\mathcal{D}(\boldsymbol p_W + \mathbf{R}\boldsymbol q_B)\leq 0, \\
		&\boldsymbol q_B =[\emph{r}\cos (o\frac{2\pi}{N_\theta}), \emph{r}\sin (o\frac{2\pi}{N_\theta}), -\frac{\emph{h}}{2}+l\frac{\emph{h}}{N_l}]^{\rm T},\\
		&\forall o \in \{0, 1,
		..., N_\theta\},
		\forall l \in \{0, 1,
		...,N_l\},\nonumber
	\end{align}
	where $\mathbf{R}$ represents the robot's orientation.
	$N_\theta$ and $N_l$ denoting user-specified angular discretization resolution and axial discretization resolution along the Z-axis, respectively. 
	Notably, our modeling approach also extends to post-grasping motion planning.  Then, the deformation radius $\emph{r}$ becomes fixed and is excluded from optimization, while the grasped object's volume integrates into the composite geometry $\mathcal{S}$.  This composite representation remains analytically differentiable with respect to optimization variables, thus facilitating gradient-based optimization. Eq. (\ref{eq:continus}) enforces $\mathcal{C}^{\widetilde{d}}$ continuity at waypoints connecting adjacent polynomial pieces. Additionally, by setting 
	$s$ to $3$, we define jerk as the measure of control effort for the trajectory.
	
	For practical solution, we leverage minimum-energy trajectory principles~\cite{wang2022geometrically} to reformulate the optimization by reparameterizing polynomial coefficients into interconnected waypoints and temporal intervals. 
	This transformation inherently satisfies equality constraints Eq. (\ref{eq:bdcon}, \ref{eq:continus}) while reducing solution space dimensionality. For handling inequality constraints Eq. (\ref{eq:va}-\ref{eq:safe}), inspired by works~\cite{han2023efficient, zhou2020ego}, we approximate continuous-time constraints through uniform sampling of points along each polynomial piece, subsequently relaxed via penalty methods. 
	Ultimately, the original problem is transformed into an analytically differentiable unconstrained optimization, efficiently solved via L-BFGS~\cite{liu1989limited}.

	\subsection{Enhanced Adaptive Control} 
	
	For flight control, we consider the quadrotor's states $\boldsymbol{x} = [\boldsymbol{p}^{\rm T}_W \  \boldsymbol{v}^{\rm T}_W \  \boldsymbol{q}^{\rm T}_W \  \bm{\omega}^{\rm T}_B]^{\rm T}$ and system inputs $\boldsymbol{u} = [F_u \  \boldsymbol{\tau}_u^{\rm T}]^{\rm T}$, and design the NMPC problem as follows:
	\begin{equation}
		\begin{aligned} 
			\boldsymbol{u}_\text{NMPC} =\mathop{\arg\min}\limits_{\boldsymbol{u}} \sum_{i=0}^{N-1}  ((\boldsymbol{x}_i-\boldsymbol{x}_{i,r})^T\boldsymbol{Q}(\boldsymbol{x}_i-\boldsymbol{x}_{i,r})  \hfill \\
			+(\boldsymbol{u}_i-\boldsymbol{u}_{i,r})^T\boldsymbol{W}(\boldsymbol{u}_i-\boldsymbol{u}_{i,r}))  \\
			+(\boldsymbol{x}_N-\boldsymbol{x}_{N,r})^T\boldsymbol{Q}_N(\boldsymbol{x}_N-\boldsymbol{x}_{N,r}), \\
			s.t.\ \  \ \  \ \  \boldsymbol{x}_{k+1} = f(\boldsymbol{x}_k, \boldsymbol{u}_k), \boldsymbol{x}_0 = \boldsymbol{x}_{now}, \ \ \ \ \ \  \\
			\boldsymbol{u} \in [\boldsymbol{u}_{min},\boldsymbol{u}_{max}], \ \ \ \ \ \ \ \  \ \ \ \ \ \  \ \ \ \ \ \ 
		\end{aligned} 
	\end{equation}
	where $N$ denotes the total time step, $i$ denotes the current time step; $\boldsymbol{x}_{i,r}$ and $\boldsymbol{x}_{N,r}$ are the reference state vectors; $\boldsymbol{u}_{i,r}$ is the reference input vector; $\boldsymbol{Q} = \text{diag}(\boldsymbol{Q}_p, \boldsymbol{Q}_v, \boldsymbol{Q}_q, \boldsymbol{Q}{\omega})$, $\boldsymbol{Q}_N$ and $\boldsymbol{W}$ are positive definite weight matrices; $\boldsymbol{u}_{min}$ and $\boldsymbol{u}_{max}$ represent the minimum and maximum thrust provided by motors, ensuring the required thrust inputs remain within feasible limits. By solving the aforementioned control optimization problem, we obtain the system's control inputs $[F_u \  \boldsymbol{\tau}_u^{\rm T}]^{\rm T}$.
	
	\subsubsection{Lightweight Force Compensation}
	
	In real-world scenarios, RR is subject to deformation-induced motion disturbances, external forces, and propeller aerodynamic effects. To enhance control performance, we consider incorporating these uncertainties into the state-space representation of the system. Based on the translational dynamics of the system, the external force disturbances can be expressed as follows: 
	\begin{equation}
		\begin{aligned}
			\boldsymbol{F}_{ext} = m\dot{\boldsymbol{v}}_W - m\boldsymbol{g} -F\boldsymbol{z}_B.			
			\label{eq: external force_est}
		\end{aligned}
	\end{equation}
	By compensating for this external disturbance in the thrust control, the desired thrust can be formulated as:
	\begin{equation}
		\begin{aligned}
			F_{des} =  || F_u\boldsymbol{z}_B -\boldsymbol{F}_{ext} ||.			
			\label{eq: compensate force_est}
		\end{aligned}
	\end{equation}

	\begin{figure}[t]
		\centering
		\includegraphics[width=1.00\linewidth]{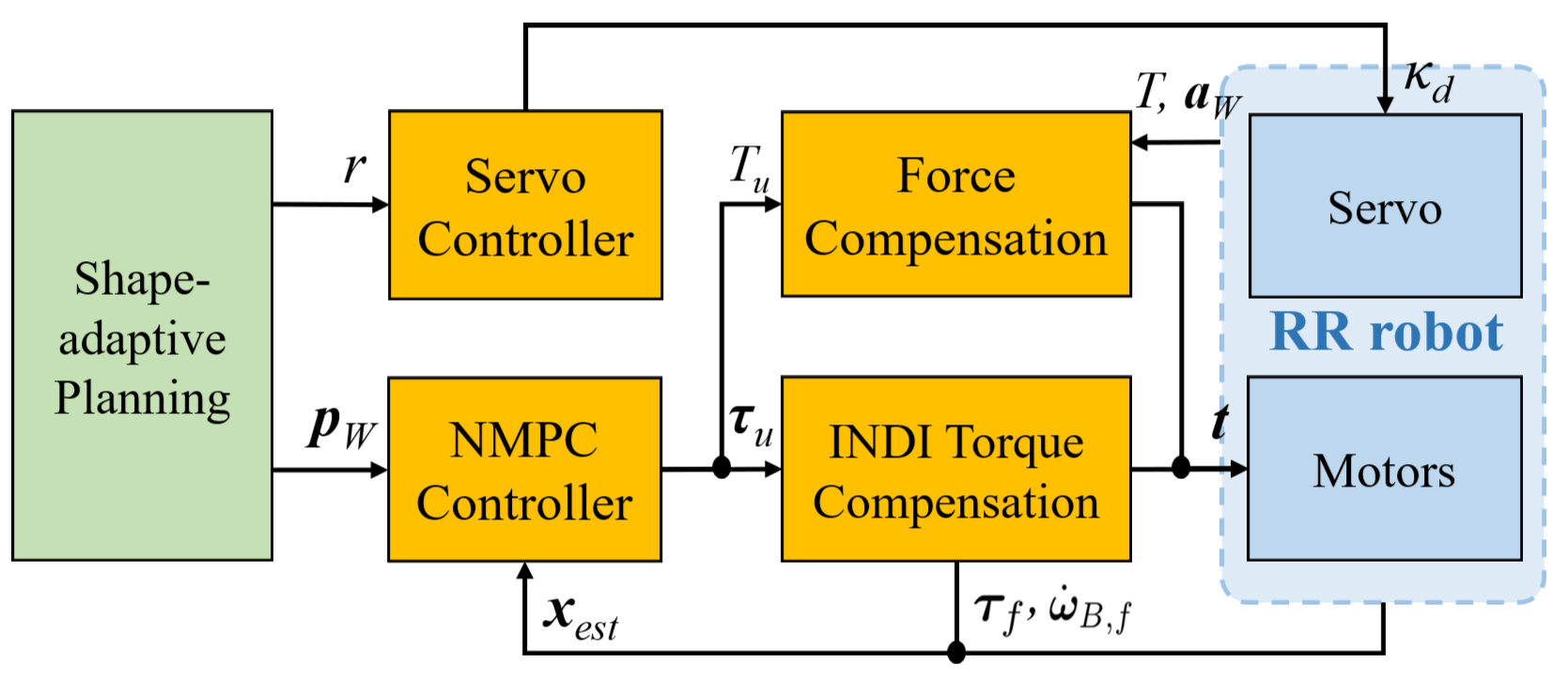}
		\captionsetup{font={small}}
		\caption{The schematic diagram of the proposed enhanced adaptive controller.}
		\label{pic:controller}
		\vspace{-0.0cm}
	\end{figure}

	\subsubsection{INDI Torque Compensation}
	We also employ an INDI controller to compensate for external torques. This strategy provides fast response to input commands and exhibits robustness to model uncertainties and external disturbances \cite{tal2020accurate,sun2022comparative}. We design the INDI control command $\boldsymbol{\tau}_\text{INDI}$ as follows:
	\begin{equation}
		\begin{aligned}
			\dot{\bm{\omega}}_{B,d} & = \boldsymbol{J}^{-1}(\bm{\tau}_u - \bm{\omega}_B\times\boldsymbol{J}\bm{\omega}_B)  \\
			\boldsymbol{\tau}_\text{INDI} & = \boldsymbol{\tau}_f + \boldsymbol{J} ( \boldsymbol{\dot{\omega}}_{B,d} - \dot{\boldsymbol{\omega}}_{B,f}),
			\label{eq:desired torque}
		\end{aligned}
	\end{equation} 
	where $\boldsymbol{\tau}_f$ is the estimated control torque in the body frame. The terms $\dot{\boldsymbol{\omega}}_{B,d}$ and $\dot{\boldsymbol{\omega}}_{B,f}$ denote the desired and estimated angular acceleration, and J is estimated and updated in real time during deformation.
	The schematic diagram of the proposed controller is illustrated in Fig.~\ref{pic:controller}. Finally, the desired thrusts of four rotors are calculated as follows:
	\begin{equation}
		\begin{aligned}
			\boldsymbol{t} =       
			{\boldsymbol{H}_k}^{-1}\begin{bmatrix}
				F_{des} \\  
				\boldsymbol{\tau}_\text{INDI}
			\end{bmatrix},	
			\label{eq: final_input}
		\end{aligned}
	\end{equation}

	\subsection{Servo Controller for Morphing} 
	\label{sec:Servo Controller}
	The servo system can be approximated as a first-order system with time constant $\gamma$. We design a proportional controller as follows: 
	\begin{equation}
		\begin{aligned} 
			{\kappa}_{d} &= \frac{1}{\gamma}(\rho(r)-\theta_{est}),
		\end{aligned} 
	\end{equation}
	where $\rho(r)$ indicates the servo motor's angle mapped to RR's radius, $\theta_{est}$ is the estimated angle, and  $\kappa_{d}$ is the desired rotating speed of the motor.

	\begin{table}[t]
		\centering	
		\caption{A trajectory comparison among the max size, min size, and our shape-adaptive planning approach in simulation.}	
		\setlength{\tabcolsep}{0.88mm}	
		\renewcommand\arraystretch{1.5}	
		{		
			\begin{tabular}{ccccccc}
				\hline
				Method   & \begin{tabular}[c]{@{}c@{}}PCE \\ ($m^2/s^5$)\end{tabular} & \begin{tabular}[c]{@{}c@{}}RCE \\ ($m^2/s^5$)\end{tabular} & SORR  & \begin{tabular}[c]{@{}c@{}}Time Cost \\ ($s$)\end{tabular} & Total Cost  & \begin{tabular}[c]{@{}c@{}}Total Energy \\ ($kj$)\end{tabular}  \\ \hline
				Ours     & 7.34                                                                   & 0.16                                                                   & 1.76  & 23.58                                                    & \textbf{32.84}     & \textbf{18.97}    \\ \hline
				Max-Size & 10.32                                                                  & 0                                                                      & 0     & 28.66                                                    & 38.98     & 22.71  \\ \hline
				Min-Size & 6.98                                                                   & 0                                                                      & 10.48 & 23.43                                                    & 41.25     & 20.60   \\ \hline
			\end{tabular}
		}	
		\label{tab:planning_benchmark}	
		\vspace{-1.4cm}	
	\end{table}

	\section{EXPERIMENTS} 
	\label{sec:experiments}
	
	\subsection{Experiment Platform} 
	\label{sec:Experiment_Platform}
	
	\begin{figure*}[t]
		\centering
		\includegraphics[width=0.95\linewidth]{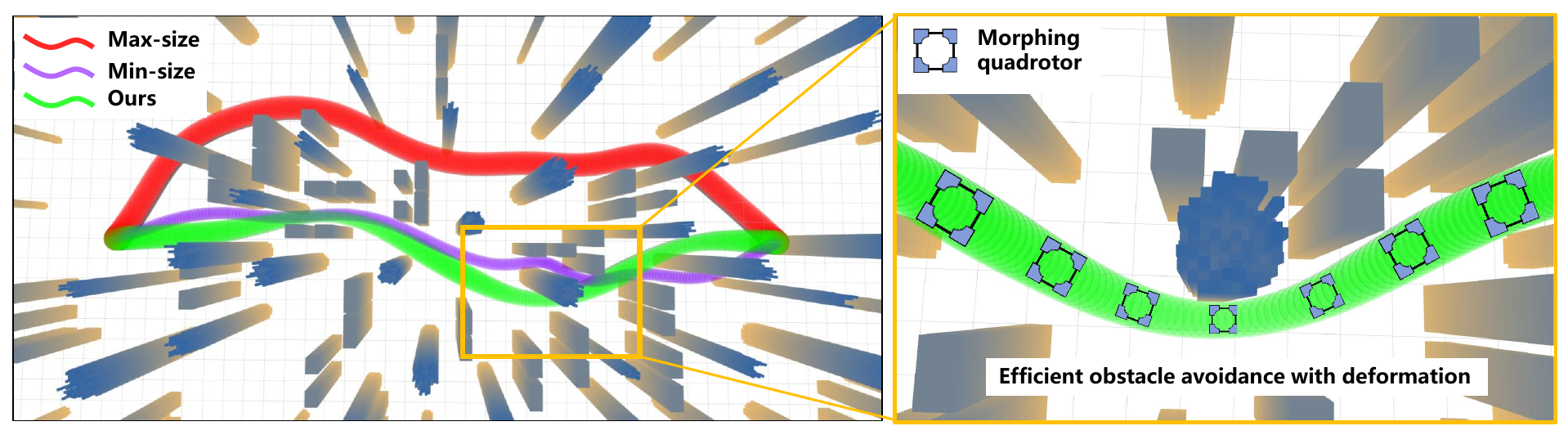}
		\captionsetup{font={small}}
		\caption{A comparison of trajectories between the proposed method and traditional fixed-size motion planning approaches~\cite{wang2022geometrically}.}
		\label{pic:comparison_planning}
		\vspace{-0.1cm}
	\end{figure*}	
	
	\begin{figure*}[htp]
		\centering
		\includegraphics[width=0.95\linewidth]{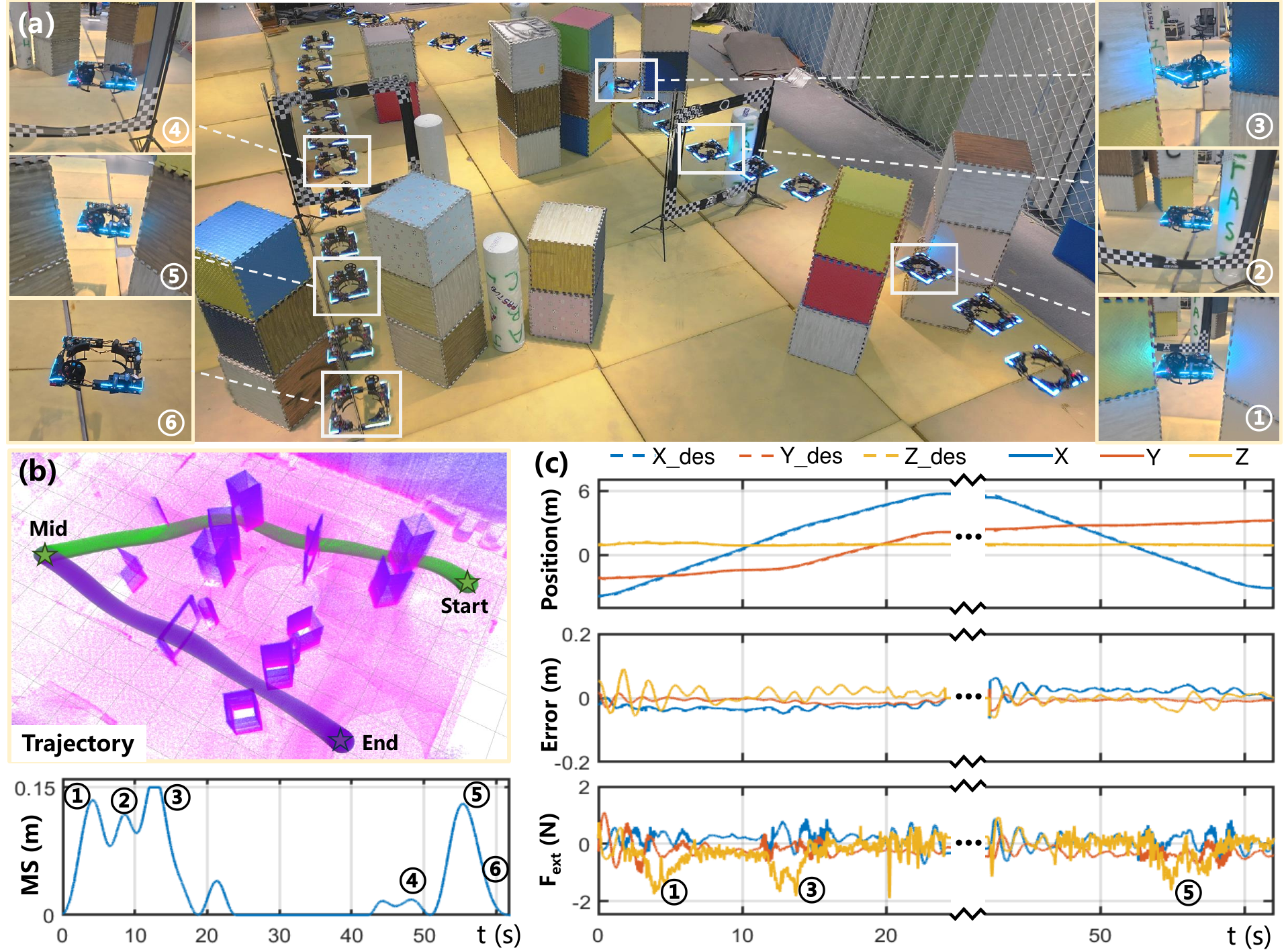}
		\captionsetup{font={small}}
		\caption{The experiment of autonomously navigating narrow and complex spaces with deformation.}
		\label{pic:morphing}
		\vspace{-0.5cm}
	\end{figure*}	
	
	We conduct simulations and real-world experiments to validate the effectiveness of the proposed algorithms. The simulations are all conducted in a Linux environment on a computer equipped with an Intel Core i7-10700 CPU. The real-world experiments utilize our optimized morphing and grasping integrated quadrotor as the experimental platform. In addition to the planning and control modules, we use a radar to construct an environmental point cloud map as environmental input, and fuse data from a NOKOV Motion Capture System and an IMU to obtain the motion states. 

	\subsection{Simulation Experiments}
	In simulations, we construct a complex environment with narrow gaps and dense obstacles to evaluate the effectiveness of our shape-adaptive motion planning, as depicted in Fig.~\ref{pic:comparison_planning}, and compare our method against traditional trajectory planning algorithms~\cite{wang2022geometrically} using fixed maximum and minimum sizes to demonstrate its advantages.
	Moreover, we calculate the total energy consumption of flight based on the power model~\cite{hoffmann2007quadrotor, wu2023ring}, which is in the form of the time integral of $||F||^{\frac{3}{2}}$ and $(\frac{\emph{r}(t)-\emph{r}_{\max}}{\emph{r}_{\max}})^2$.
	Furthermore, the average metric costs from hundreds of planning runs are summarized in Tab.~\ref{tab:planning_benchmark}, with the meaning of each abbreviation for these metrics detailed in Sect.~\ref{sec:EA_planning} under the trajectory optimization problem formulation Eq. (\ref{eq:opt}).
	The data shows that the max-size planning cannot leverage the passability of shape adjustment, lagging behind in both time, energy and trajectory position control effort. 
	While the fixed min-size planning maximizes passage performance, sustained flight in the smallest configuration increases energy consumption (Tab.~\ref{tab:planning_benchmark}), and is not conducive to practical flight performance~\cite{2019Falanga}. 
	In contrast, our shape-adaptive trajectory planning dynamically adjusts the robot's dimensions based on environmental demands, reducing size only when necessary for obstacle avoidance (Fig.~\ref{pic:comparison_planning}). This adaptive strategy achieves a favorable equilibrium across various cost metrics, minimizing time consumption close to the min size while maintaining a larger size as much as possible to reduce energy consumption (Tab.~\ref{tab:planning_benchmark}). As a result, it demonstrates significantly superior performance in terms of total cost.
	
	\begin{figure*}[htp]
		\centering
		\includegraphics[width=0.91\linewidth]{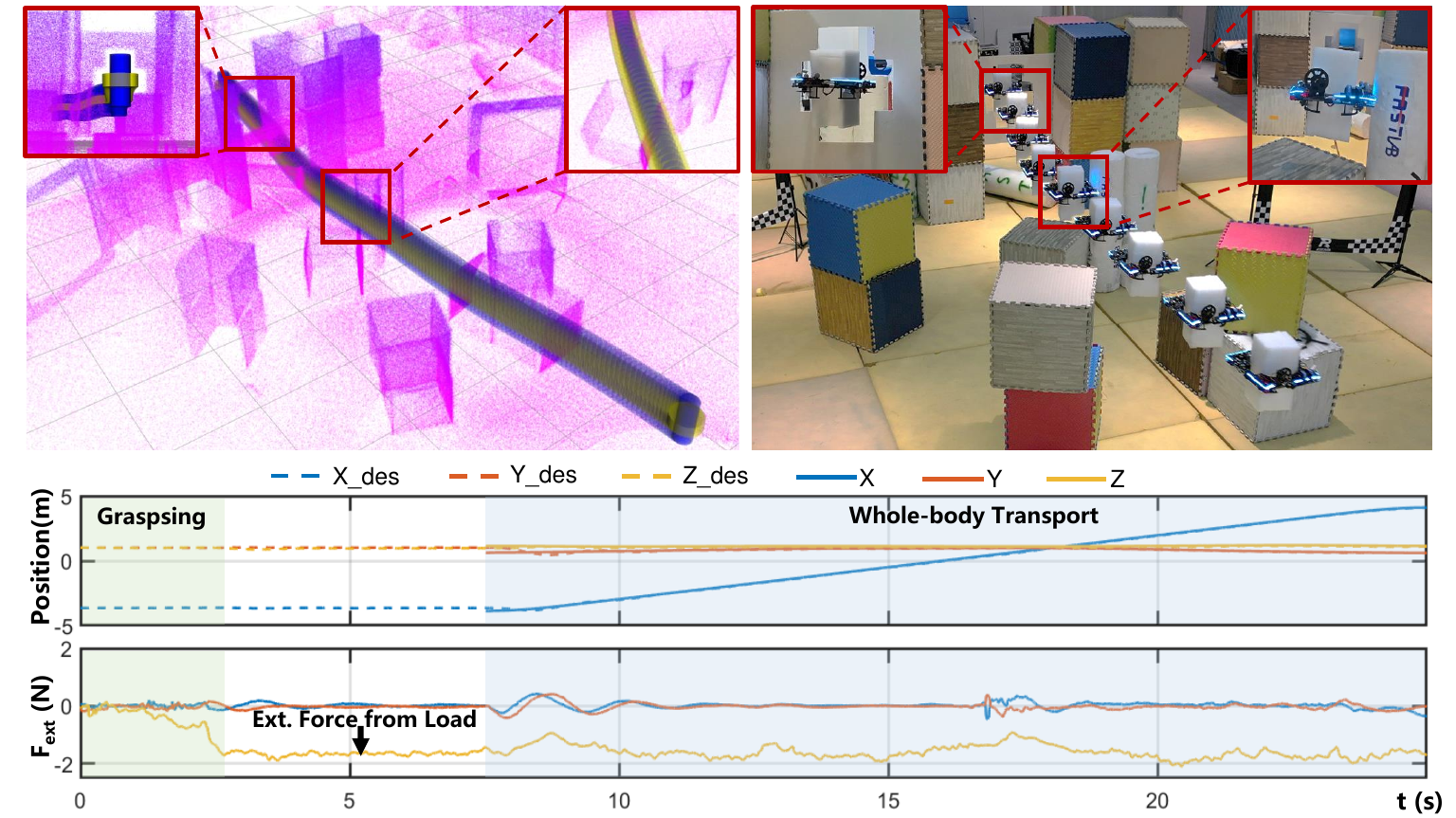}
		\captionsetup{font={small}}
		\caption{The autonomous experiment on whole-body grasping-transport with large objects through irregular environments.}
		\label{pic:transport}
		\vspace{-0.8cm}
	\end{figure*}

	\subsection{Real-world Experiments}
	
	\textbf{\emph{1) Deformation Navigation}}: we also validate the effectiveness of the shape-adaptive planning algorithm in real-world experiments. We set up a complex environment containing multiple narrow gaps (40 cm, $<$ RR's size) and densely cluttered obstacle regions, which poses a significant challenge for autonomous morphing traversal. As shown in Fig.~\ref{pic:morphing}(b), RR can generate a variable-scale trajectory and a sequence of morphing states (MS) that optimize the comprehensive cost, based on the constructed point cloud map. As shown in Fig.~\ref{pic:morphing}(a), RR sequentially undergoes five morphing transformations, which not only enable it to achieve greater passability through narrow gaps but also allow it to avoid obstacles more efficiently with shorter paths. The proposed enhanced controller estimates external disturbance forces, such as deformation disturbances, ultimately achieving an average control error of less than 5 cm (Fig.~\ref{pic:morphing}(c)).

	\textbf{\emph{2) Whole-body Grasping-Transport}}: RR's unique deformation characteristics not only enable efficient obstacle avoidance during flight but also allow for whole-body grasping-transport. Importantly, our planner (as described in Sect.~\ref{sec:EA_planning}) can be easily extended to incorporate the grasped object into the obstacle avoidance calculations, generating full-shape trajectories considering the object's presence. To validate this capability, we design a complex scenario with a narrow cross-shaped gap (only 8 cm wider than RR's size with object) and multiple obstacles. As shown in Fig.~\ref{pic:transport}, after grasping an object measuring 20 cm in length and width, 40 cm in height, and weighing 160 g, the motion planner successfully generates a grasping-transport trajectory that traverses the unstructured gap. This scenario also places high demands on the robot's precise control. Through the enhanced controller, RR accurately estimates the external forces mainly caused by the payload, ultimately successfully leveraging its grasping-transport capabilities to carry the object through the obstacle region.
	
	\begin{figure}[t]
		\centering
		\includegraphics[width=0.95\linewidth]{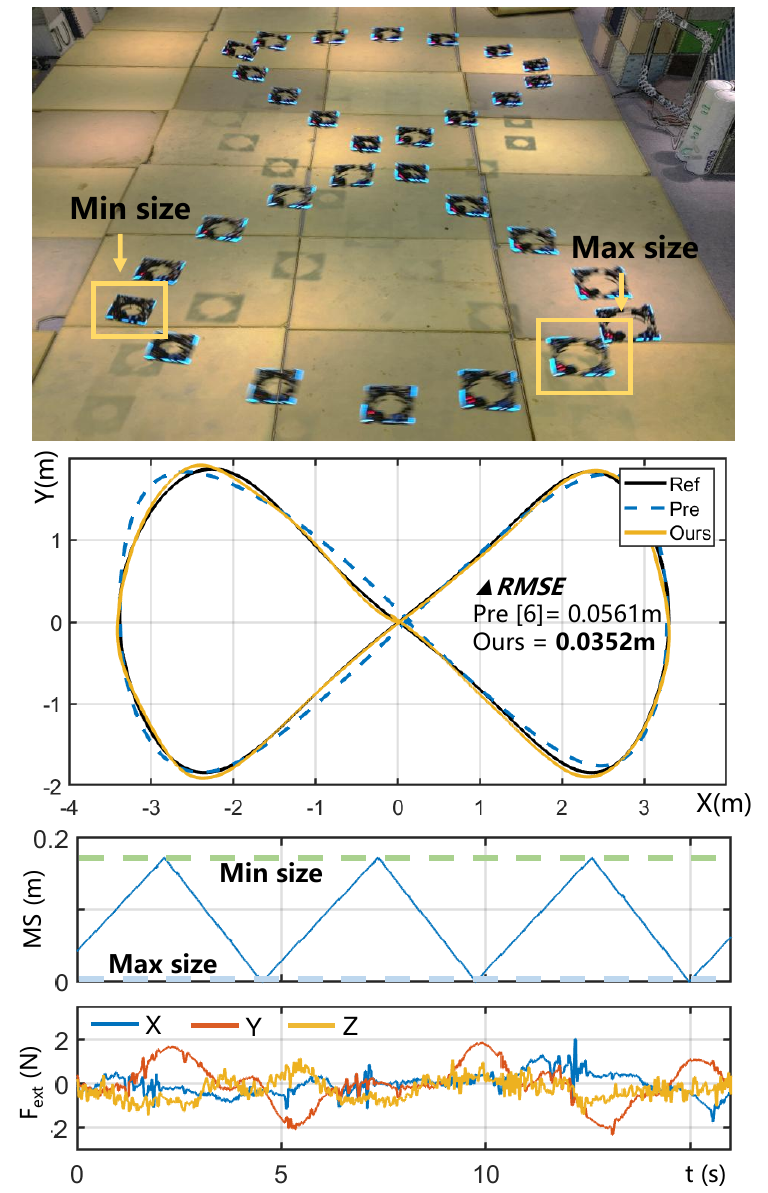}
		\captionsetup{font={small}}
		\caption{Enhanced control performance in tracking 8-figure trajectories with morphing, and comparison with the previous method \cite{wu2023ring}.}
		\label{pic:benchmark_8_2d}
		\vspace{-0.4cm}
	\end{figure}

	\textbf{\emph{3) Enhanced Control Performance}}: to evaluate the performance of the enhanced adaptive controller, we conduct experiments on deforming and tracking a figure-eight trajectory. We also compare its performance against our previously developed controller \cite{wu2023ring}. With a maximum reference trajectory speed of 1.5 m/s, the tracking results, shown in Fig.~\ref{pic:benchmark_8_2d}, demonstrate that the enhanced controller achieves more accurate tracking. Data analysis shows that the root mean square error (RMSE) of the enhanced controller is 0.0352 m, which is a 37.3\% reduction compared to the previous controller. This improvement stems from the implementation of a force and torque compensation strategy that estimates external disturbances in real-time (Fig.~\ref{pic:benchmark_8_2d}). This strategy effectively addresses model mismatches, deformation disturbances, and variations in thrust coefficients during the morphing process. The above mentioned deformation and grasping-transport experiments have also demonstrated the effectiveness of this approach in multimodal deformation tasks.

	\section{CONCLUSIONS}
	
	This paper presents a shape-adaptive motion planning and control framework for a deformable quadrotor, enabling autonomous navigation in complex confined environments. Simulations and experiments validate the approach in narrow-gap traversal and grasping-transport. These results demonstrate the potential of deformable drones in challenging autonomous scenarios. Future work will focus on handling dynamic environments with moving obstacles, exploring learning-based methods to improve robustness under uncertainties, and integrating advanced sensing like 3D mapping and object recognition for enhanced autonomy. 
	
	\addtolength{\textheight}{-12cm}   
	
	%
	%

	\bibliography{IROS2025_final}

\begin{thebibliography}{10}
\providecommand{\url}[1]{#1}
\csname url@rmstyle\endcsname
\providecommand{\newblock}{\relax}
\providecommand{\bibinfo}[2]{#2}
\providecommand\BIBentrySTDinterwordspacing{\spaceskip=0pt\relax}
\providecommand\BIBentryALTinterwordstretchfactor{4}
\providecommand\BIBentryALTinterwordspacing{\spaceskip=\fontdimen2\font plus
\BIBentryALTinterwordstretchfactor\fontdimen3\font minus \fontdimen4\font\relax}
\providecommand\BIBforeignlanguage[2]{{%
\expandafter\ifx\csname l@#1\endcsname\relax
\typeout{** WARNING: IEEEtran.bst: No hyphenation pattern has been}%
\typeout{** loaded for the language `#1'. Using the pattern for}%
\typeout{** the default language instead.}%
\else
\language=\csname l@#1\endcsname
\fi
#2}}

\bibitem{2019Falanga}
D.~Falanga, K.~Kleber, S.~Mintchev, D.~Floreano, and D.~Scaramuzza, ``The foldable drone: A morphing quadrotor that can squeeze and fly,'' \emph{IEEE Robotics and Automation Letters}, vol.~4, no.~2, pp. 209--216, 2019.

\bibitem{2017Desbiez}
A.~Desbiez, F.~Expert, M.~Boyron, J.~Diperi, S.~Viollet, and F.~Ruffier, ``X-morf: A crash-separable quadrotor that morfs its x-geometry in flight,'' in \emph{2017 Workshop on Research, Education and Development of Unmanned Aerial Systems (RED-UAS)}, 2017, pp. 222--227.

\bibitem{2018Moju}
M.~Zhao, T.~Anzai, F.~Shi, X.~Chen, K.~Okada, and M.~Inaba, ``Design, modeling, and control of an aerial robot dragon: A dual-rotor-embedded multilink robot with the ability of multi-degree-of-freedom aerial transformation,'' \emph{IEEE Robotics and Automation Letters}, vol.~3, no.~2, pp. 1176--1183, 2018.

\bibitem{2017Moju}
M.~Zhao, K.~Kawasaki, X.~Chen, S.~Noda, K.~Okada, and M.~Inaba, ``Whole-body aerial manipulation by transformable multirotor with two-dimensional multilinks,'' in \emph{2017 IEEE International Conference on Robotics and Automation (ICRA)}, 2017, pp. 5175--5182.

\bibitem{Bucki2021DesignAC}
N.~Bucki, J.~Tang, and M.~W. Mueller, ``Design and control of a midair reconfigurable quadcopter using unactuated hinges,'' \emph{ArXiv}, vol. abs/2103.16632, 2021.

\bibitem{wu2023ring}
Y.~Wu, F.~Yang, Z.~Wang, K.~Wang, Y.~Cao, C.~Xu, and F.~Gao, ``Ring-rotor: A novel retractable ring-shaped quadrotor with aerial grasping and transportation capability,'' \emph{IEEE Robotics and Automation Letters}, vol.~8, no.~4, pp. 2126--2133, 2023.

\bibitem{tordesillas2021mader}
J.~Tordesillas and J.~P. How, ``Mader: Trajectory planner in multiagent and dynamic environments,'' \emph{IEEE Transactions on Robotics}, vol.~38, no.~1, pp. 463--476, 2021.

\bibitem{zhang2023model}
R.~Zhang, J.~Lin, Y.~Wu, Y.~Gao, C.~Wang, C.~Xu, Y.~Cao, and F.~Gao, ``Model-based planning and control for terrestrial-aerial bimodal vehicles with passive wheels,'' in \emph{2023 IEEE/RSJ International Conference on Intelligent Robots and Systems (IROS)}.\hskip 1em plus 0.5em minus 0.4em\relax IEEE, 2023, pp. 1070--1077.

\bibitem{ren2022bubble}
Y.~Ren, F.~Zhu, W.~Liu, Z.~Wang, Y.~Lin, F.~Gao, and F.~Zhang, ``Bubble planner: Planning high-speed smooth quadrotor trajectories using receding corridors,'' in \emph{2022 IEEE/RSJ International Conference on Intelligent Robots and Systems (IROS)}.\hskip 1em plus 0.5em minus 0.4em\relax IEEE, 2022, pp. 6332--6339.

\bibitem{han2021fast}
Z.~Han, Z.~Wang, N.~Pan, Y.~Lin, C.~Xu, and F.~Gao, ``Fast-racing: An open-source strong baseline for {SE(3)} planning in autonomous drone racing,'' \emph{IEEE Robotics and Automation Letters}, vol.~6, no.~4, pp. 8631--8638, 2021.

\bibitem{wu2024whole}
T.~Wu, Y.~Chen, T.~Chen, G.~Zhao, and F.~Gao, ``Whole-body control through narrow gaps from pixels to action,'' \emph{arXiv preprint arXiv:2409.00895}, 2024.

\bibitem{cui2024motion}
G.~Cui, R.~Xia, X.~Jin, and Y.~Tang, ``Motion planning and control of a morphing quadrotor in restricted scenarios,'' \emph{IEEE Robotics and Automation Letters}, 2024.

\bibitem{derrouaoui2021pso}
S.~H. Derrouaoui, Y.~Bouzid, and M.~Guiatni, ``Pso based optimal gain scheduling backstepping flight controller design for a transformable quadrotor,'' \emph{Journal of Intelligent \& Robotic Systems}, vol. 102, no.~3, p.~67, 2021.

\bibitem{kim2021morphing}
C.~Kim, H.~Lee, M.~Jeong, and H.~Myung, ``A morphing quadrotor that can optimize morphology for transportation,'' in \emph{2021 IEEE/RSJ International Conference on Intelligent Robots and Systems (IROS)}.\hskip 1em plus 0.5em minus 0.4em\relax IEEE, 2021, pp. 9683--9689.

\bibitem{2021Hu}
D.~Hu, Z.~Pei, J.~Shi, and Z.~Tang, ``Design, modeling and control of a novel morphing quadrotor,'' \emph{IEEE Robotics and Automation Letters}, vol.~6, no.~4, pp. 8013--8020, 2021.

\bibitem{zhou2019robust}
B.~Zhou, F.~Gao, L.~Wang, C.~Liu, and S.~Shen, ``Robust and efficient quadrotor trajectory generation for fast autonomous flight,'' \emph{IEEE Robotics and Automation Letters}, vol.~4, no.~4, pp. 3529--3536, 2019.

\bibitem{wang2022geometrically}
Z.~Wang, X.~Zhou, C.~Xu, and F.~Gao, ``Geometrically constrained trajectory optimization for multicopters,'' \emph{IEEE Transactions on Robotics}, vol.~38, no.~5, pp. 3259--3278, 2022.

\bibitem{han2023efficient}
Z.~Han, Y.~Wu, T.~Li, L.~Zhang, L.~Pei, L.~Xu, C.~Li, C.~Ma, C.~Xu, S.~Shen, \emph{et~al.}, ``An efficient spatial-temporal trajectory planner for autonomous vehicles in unstructured environments,'' \emph{IEEE Transactions on Intelligent Transportation Systems}, vol.~25, no.~2, pp. 1797--1814, 2023.

\bibitem{zhou2020ego}
X.~Zhou, Z.~Wang, H.~Ye, C.~Xu, and F.~Gao, ``Ego-planner: An esdf-free gradient-based local planner for quadrotors,'' \emph{IEEE Robotics and Automation Letters}, vol.~6, no.~2, pp. 478--485, 2020.

\bibitem{liu1989limited}
D.~C. Liu and J.~Nocedal, ``On the limited memory bfgs method for large scale optimization,'' \emph{Mathematical programming}, vol.~45, no.~1, pp. 503--528, 1989.

\bibitem{tal2020accurate}
E.~Tal and S.~Karaman, ``Accurate tracking of aggressive quadrotor trajectories using incremental nonlinear dynamic inversion and differential flatness,'' \emph{IEEE Transactions on Control Systems Technology}, vol.~29, no.~3, pp. 1203--1218, 2020.

\bibitem{sun2022comparative}
S.~Sun, A.~Romero, P.~Foehn, E.~Kaufmann, and D.~Scaramuzza, ``A comparative study of nonlinear mpc and differential-flatness-based control for quadrotor agile flight,'' \emph{IEEE Transactions on Robotics}, vol.~38, no.~6, pp. 3357--3373, 2022.

\bibitem{hoffmann2007quadrotor}
G.~Hoffmann, H.~Huang, S.~Waslander, and C.~Tomlin, ``Quadrotor helicopter flight dynamics and control: Theory and experiment,'' in \emph{AIAA guidance, navigation and control conference and exhibit}, 2007, p. 6461.

\end{thebibliography}
\end{document}